\documentclass[10pt,twocolumn,letterpaper]{article}

\usepackage{cvpr}
\usepackage{times}
\usepackage{epsfig}
\usepackage{graphicx}
\usepackage{amsmath}
\usepackage{amssymb}

\usepackage[pagebackref=true,breaklinks=true,letterpaper=true,colorlinks,bookmarks=false]{hyperref}

\cvprfinalcopy 

\def\cvprPaperID{****} 

\ifcvprfinal\fi
\begin{document}

\title{ Curvature: A signature for Action Recognition in Video Sequences}

\author{He Chen\\
Johns Hopkins University\\
{\tt\small hchen136@jhu.edu}
\and
Gregory S. Chirikjian\\
National University of Singapore, Johns Hopkins University\\
{\tt\small  gchirik1@jhu.edu}
}

\maketitle

\begin{abstract}
   In this paper, a novel signature of human action recognition, namely the curvature of a video sequence, is introduced. In this way, the distribution of sequential data is modeled, which enables few-shot learning. Instead of depending on recognizing features within images, our algorithm views actions as sequences on the universal time scale across a whole sequence of images. The video sequence, viewed as a curve in pixel space, is aligned by reparameterization using the arclength of the curve in pixel space. Once such curvatures are obtained, statistical indexes are extracted and fed into a learning-based classifier. Overall, our method is simple but powerful. Preliminary experimental results show that our method is effective and achieves state-of-the-art performance in video-based human action recognition. Moreover, we see latent capacity in transferring this idea into other sequence-based recognition applications such as speech recognition, machine translation, and text generation.
\end{abstract}

\section{Introduction}

Action Recognition based on AI-reasoning is one of the most important research topics in computer vision\cite{Authors01,Authors02,Authors03,Authors04}. In recent years, this field has witnessed big breakthroughs, and the research interest is evolving from learning the joints of human and recognizing human pose to the understanding of actions and scenes.\\ 
\indent With the booming of learning-based AI technology, precision of action recognition has been raised to a new level\cite{Authors05}. However, challenges still exist in this field. Most of these learning-based algorithms are thirsty for large datasets with tens of thousands of videos and corresponding labels. Although thousands of videos are uploaded to Youtube every second, they are raw data which can't be fed directly into most of existing algorithms. The most time-consuming part is getting detailed, fine-tuned annotations, which usually involves some manual efforts\cite{Authors06,Authors07}.\\
\indent Encoding action into a model is important because this might lead to effective extraction of motion features, thus effectively decrease the amount of training data required. Modeling of human action could be categorized into two perspectives, namely spatially and temporally. Spatially, the shape of human could be modeled into several forms such as skeleton\cite{Authors08}, silhouettes\cite{Authors09}, virtual skin\cite{Authors010}, etc. Temporally, action could be modeled into video sequence, audio sequence, dynamic image\cite{Authors011}, etc. In this paper, we focus on the second perspective.\\
\indent In many application scenarios, actions or scenes need to be understood on universal time scale. For example, let's consider the scenario of service robot in a house as shown in Fig.1(a). By capturing a single image, the robot could be confused whether the lady wants to put down or pick up the cup. What if we capture six consecutive frames and see that the hand is moving towards table? Does she want to put the cup onto the table? Still hard to tell. It could be an old lady with trembling hand trying to pick up the cup. Locally, the cup could be approaching the table, while globally, the lady might want to pick up the cup from the table.\\

\begin{figure}[thpb]
      \centering
      \includegraphics[scale=0.08]{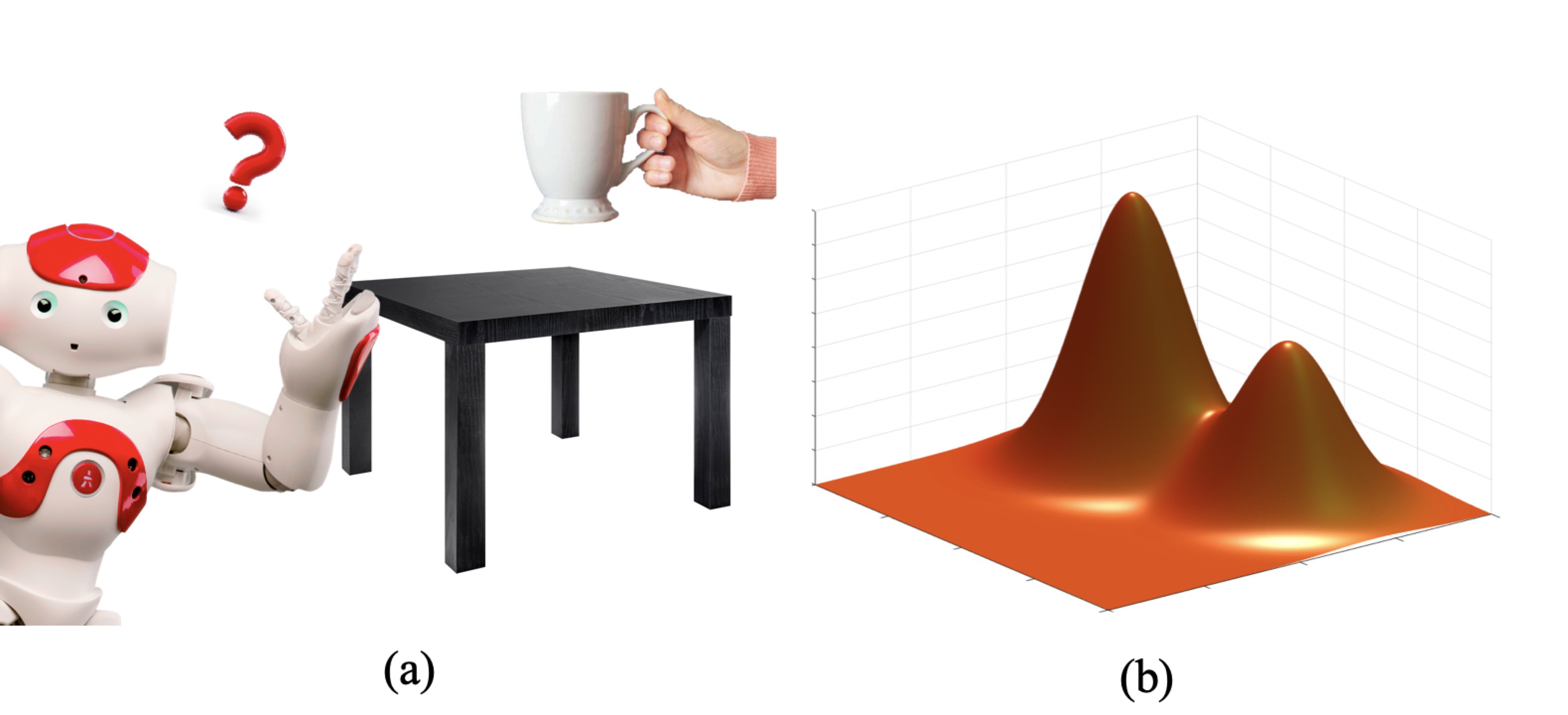}
      \caption{(a) Instance of action recognition in the application of service robots (b) Multimodal distribution}
      \label{figurelabel}
   \end{figure}

\indent From the aspect of modeling, the problem of action recognition boils down to decreasing the distance between the distributions actions from the same class, while at the same time increasing the distance between that of different classes on universal time scale.\\
\indent In order to get a better understanding of sequences on a global timescale, \cite{Authors012} proposed the theory of globally optimal reparameterization algorithm (GORA) using variational calculus. Consider the example in \cite{Authors012}, when $X_{i}(t)$ are two scalar functions each describing the audio signal of spoken text 'The rain in Spain stays mainly in the plain.' $X_{1}(t)$ could be the template of how this phrase should be spoken, and $X_{2}(t)$ could be how someone with an accent says (or sings) the same phrase. That's to say we need to find a signature of sequence which eliminates noises by modeling. The same is true when it comes to action recognition.\\
\indent In this paper, we propose a novel few-shot learning algorithm using curvature as a signature for action recognition. In this algorithm, we use the curvature of an optimally reparameterized video sequence as a signature of action. Such a model is based on viewing a video sequence as a curve in the pixel space, the curve parameters defining the speed of traversal through the sequence. The proposed algorithm considers and analyzes action from the global time scale, which is the arclength of the curvature, and expresses actions in a very compact way. After curvatures are obtained, we gather important features from these curvatures and input them into the classifier. Random forest is selected as the classifier. And it should be noted that our algorithm is robust against flipping, which is to say walking from left to right has the same signature as walking from right to left. This algorithm is also invariant to mirror reflections and rigid-body displacements of image plane. Moreover, because we modeled the distribution of data with an effective and compact model, our algorithm is not thirsty for large-scale data. In order to eliminate the noise of background, we used state-of-the-art segmentation algorithm Mask R-CNN as a pre-processing. Experimental results show that the proposed algorithm is effective and has good performance.

\section{Related Work}
\indent In this section, a brief review is given for existing state-of-the-art methods in the field of action classification and representation. Due to limitation of space, we could not list all of them, but please refer to the reviews\cite{Authors013, Authors014} if you have further interest. 
\subsection{Action Classification Based on RGB Videos}
In the early stage of the development of action recognition, histogram of gradient (HOG) and histogram of flow (HOG) algorithms are widely accepted structures. The main idea of HOG\cite{Authors015} is dividing images into small cells, and draw the histogram of the gradients of edges for these cells. While HOF\cite{Authors016} is based on measuring the angle between optical flow vector and x-axis. In\cite{Authors017, Authors018}, H.Wang et al. proposed dense trajectory as a description of videos to accomplish action classification task, and improved it to be more robust against camera motions with optical flow. Thanks to the development of computer structure and learning theory, recent years have witnessed significant improvements of learning-based solutions for action recognition on RGB videos. S.Ji et al.\cite{Authors019} extended conventional CNN from image classification to videos by taking the time scale into consideration. And D.Tran et al.\cite{Authors020} made further improvement on CNN based method by modeling appearance and motion simultaneously and found the best kernel for the network. In \cite{Authors021}, A.Karpathy et al. proposed Deep Video algorithm which enables CNN with the ability to classify large-scale dataset of 1 million videos with the multi-resolution architecture. Admittedly, these algorithms show excellent results on most benchmark datsets, the prerequisit of them usually includes a large training dataset. 
\subsection{Representation of Actions}
\indent In recent years, more and more attention is paid to alleviate the training burden by making improvements to modeling\cite{Authors039}. Girdhar at al. \cite{Authors022} learned attention maps instead of whole videos to make classification more efficient. In \cite{Authors023}, Guo et al. proposed a graphical representation for spatial information in 3D data using neural graph matching networks, which enables few-shot learning. Bilen et al.\cite{Authors011} proposed a novel compact representation of actions using dynamic image. In their algorithm, dynamic images are obtained by directly applying rank pooling on the raw image pixels of a video, producing a single RGB image representation for each video. In \cite{Authors024}, Zhou et al. used temporal relational reasoning to analyze the current situation relative to the past and formulate what may happen next. Long short term memory (LSTM) algorithm was introduced to cope with video sequences in the field of action recognition \cite{Authors025, Authors026}. LSTM, the well known language processing algorithm is breaking its way into the vision field, showing promising results and huge potential. Experimental results of multiple papers show that LSTM outperforms CNN-based methods when it comes to understanding action or scene. If we look into the description behind the fancy network architectures, it's kind of natural because LSTM architecture models the input as a sequence, which is closer to the physics meaning of action.

\section{Proposed Signature}
For human, the task of recognition is one of the first skills we learn from the moment we are born and is one that comes naturally and effortlessly as adults. It is commonly believed that edge-detection of objects plays important role in the eyesight of human. However, the optical system of the human eye is vastly more complicated than edge-detector. Medical research carried out by L.Riggs in literature \cite{Authors027} shows that curvature is a specific feature of human vision perception, and detectors of curvature exist in human eyes that generate color-contingent aftereffects. In \cite{Authors028}, M.Kass et.all proposed an active contour model called snake, which is also a very interesting way to describe curvature in the image plane. L.Gorman proposed in \cite{Authors029} that conversion of a image into a representation of curves and line features enable economical storage of information. \\
\indent Unlike many works, we are not focusing on curvature of contours in the image plane, but rather curvature of the video sequence, which is viewed as a curve in $d^2$ dimensional Euclidean space. In this section, we firstly give a brief review of globally optimal reparameterization algorithm (GORA), then present the generation of curvature based on GORA.
\subsection{Review of GORA}
Assume that $\textbf{\emph{X}}_{1}(t)$ and $\textbf{\emph{X}}_{2}(t)$ are two signals on the time scale $t \in [0,1]$, which might have different temporal frequencies over this range. Thus, even if $\textbf{\emph{X}}_{1}(t)$ and $\textbf{\emph{X}}_{2}(t)$ represents the same action or audio sequence, it is still possible that the distance between these two signals is not close to zero at all, which is not the ideal status we would like to see. GORA \cite{Authors012} solves this problem by deriving a monotonically increasing function $\tau(t)$ that normalize signals onto universal standard timescale. The expression for $\tau(t)$ is derived using variational calculus, whose result is
$$\tau (t) = F^{-1}(t)\eqno{(1)}$$
where $F$ has the form of
$$F(\tau) = \frac{1}{c} \int_{0}^{\tau} \bigg| \bigg|\frac{d\textbf{\emph{X}}}{dt} \bigg| \bigg| dt\eqno{(2)}$$
$$c = \int_{0}^{1} \bigg| \bigg|\frac{d\textbf{\emph{X}}}{dt} \bigg| \bigg| dt$$
Then, if $X_{i}(t) = X(s_{i}(t))$, for $i = 1,2$, where $s_{i}(t)$ are arbitrary monotonically increasing functions of time, reparameterization will recognize them as the same signal, by quotienting out the effects of $s_{i}(t)$.

\begin{figure*}
\begin{center}
\includegraphics[width=0.8\linewidth]{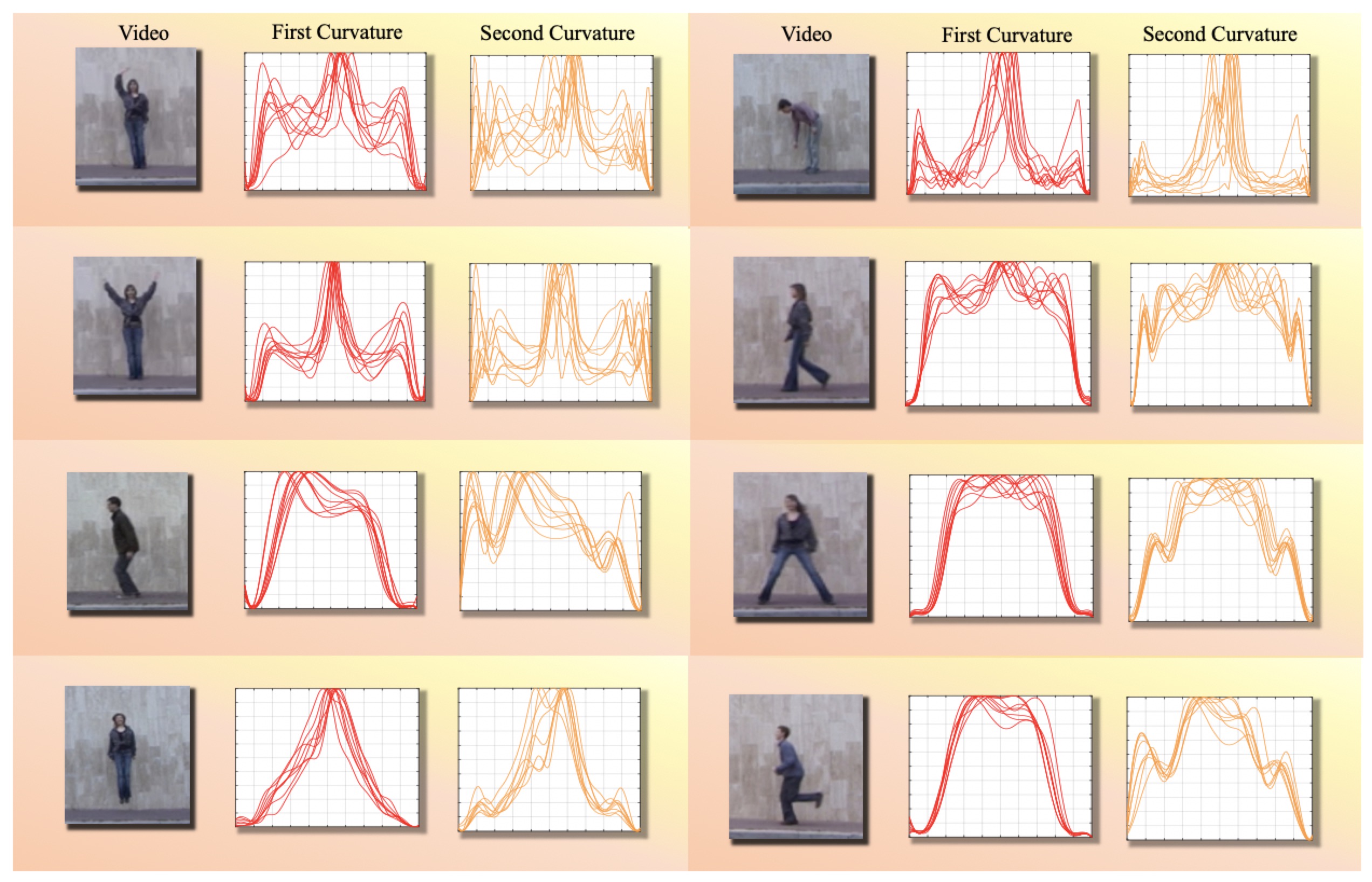}
\end{center}
   \caption{curvature as a signature}
\label{fig:short}
\end{figure*}

\subsection{Curvature Generation}
Intuitively, curvature is the amount by which a geometric object, such as a surface, deviates from being a flat plane or a curve from being straight as in the case of a line\cite{Authors030}.\\
\indent That is the narrower definition of curvature. In this paper, we mean the broader definition of curvature, which is calculated by $n$th-order derivatives of a curve in n-dimensional Euclidean space. By curvature, we are referring to the curvature of the curve using the Frenet frame \cite{Authors030}. Assume that $\textbf{\emph{X}} = \textbf {\emph{X}}(s)$ $\epsilon$ $E^{n}$ is a parametric representation of a generally curved curve with arclength $s$. Then the following derivation equations are valid
$$\frac{d\textbf{\emph{X}}}{ds} = \textbf{\emph{e}}_1\eqno{(3)}$$
$$\frac{d\textbf{\emph{e}}_{1}}{ds} = \textbf{\emph{e}}_{2}k_{1}\eqno{(4)}$$
$$\frac{d\textbf{\emph{e}}_{2}}{ds} = -\textbf{\emph{e}}_{1}k_{1} +\textbf{\emph{e}}_{3}k_{2}\eqno{(5)}$$
$$......$$
$$\frac{d\textbf{\emph{e}}_{n-1}}{ds} = -\textbf{\emph{e}}_{n-2}k_{n-2} +\textbf{\emph{e}}_{n}k_{n-1}\eqno{(6)}$$
$$\frac{d\textbf{\emph{e}}_{n-1}}{ds} = -\textbf{\emph{e}}_{n}k_{n-1}\eqno{(7)}$$
where $s = F \cdot c$ represents the arc length, $\textbf{\emph{e}}_{1}$,...,$\textbf{\emph{e}}_{n}$ represents the orthogonal basis, and $\textbf{\emph{k}}_{1}$,...,$\textbf{\emph{k}}_{n}$ represents the curvatures. $\textbf{\emph{k}}_{1}$ is the curvature that is most widely used, while $\textbf{\emph{k}}_{2}$,...,$\textbf{\emph{k}}_{n}$ are higher curvatures. Since curvature is an intrinsic quality, if a reflection, rotation, or translation is applied to the video sequence, the curvature will be invariant.

\section{Action Recognition Based on Curvature Calculation}

Now that we already compressed the patterns of actions into curvatures on the global timescale, we could use these curvatures to accomplish the action recognition task. In order to achieve that, first of all, let's see the curvatures as one-dimensional distributions with respect to global time scale. Then, we input these features into a classifier to distinguish between different actions. Here, we choose to use Random Forest, which is one of the most powerful machine-learning based classifiers has a simple structure, very easy to implement, and usually wouldn't overfit.

\subsection{Feature Selection} 
When it comes to one-dimensional distributions, most commonly applied features including 
\begin{itemize}
\item {feature of center position}: mean $\mu$, median $m$
\item {feature of divergence}: range $r$, standard deviation $\sigma$
\end{itemize}
And fancier statistical features including
\begin{itemize}
\item wave rate: ninty-percent-quantile minus ten-percent-quantile $t$
\item skewness: a measure of asymmetry about a distribution about its mean
$$Skew(X) = E \left[(\frac{X - \mu}{\sigma})^3\right]\eqno{(8)}$$
\item kurtosis: a measure of sharp or flat about a distribution
$$Kurt(X) = E\left[(\frac{X - \mu}{\sigma})^4\right]\eqno{(9)}$$
\end{itemize}
However, all aforementioned statistical features could be deceiving sometimes when distributions similar to multimodal distributions which have similar looks with Fig.1(b) are considered. That's to say, even all the aforementioned indexes are the same for two distributions, the two distributions still look very different. In order to fix this problem, we introduce another index $Beta$ \cite{Authors031}
$$Beta(X) = \frac{Skew(X)^2+1}{Kurt(X)}\eqno{(10)}$$

\subsection{Action Classification}
Random forest is a powerful classifier belonging to the family of ensemble learning, whose component is decision trees. This ensemble tree bag consists of multiple decision trees. When a classification problem is given to a random forest, each tree might have its own idea and vote, the class that gains most votes would be the result for the final decision.\\
\indent In order to generate decision trees, if the training dataset has the size of $N$ for each tree, we randomly select $N$ data from the whole dataset and put those $N$ data back to the dataset after usage. This bootstrapping mode is applied for sampling.\\
\indent Assuming that the dimension of features is $M$, and we select a constant $m << M$. A subset consisting of $m$ randomly selected element is formed to train each tree. When the tree grows, optimal features are considered. No pruning is required in the structure of random forest.\\
\indent Each decision tree is a CART (Classification And Regression Tree). The trees are trained by minimizing Gini coefficient in the form of
$$Gini(p) = \Sigma p_{k}(1 - p_{k})\eqno{(11)}$$
where $k$ is the number of classes, and the probability that a sampled point belongs to class $k$ is $p_{k}$.

\begin{figure*}
\begin{center}
\includegraphics[width=0.9\linewidth]{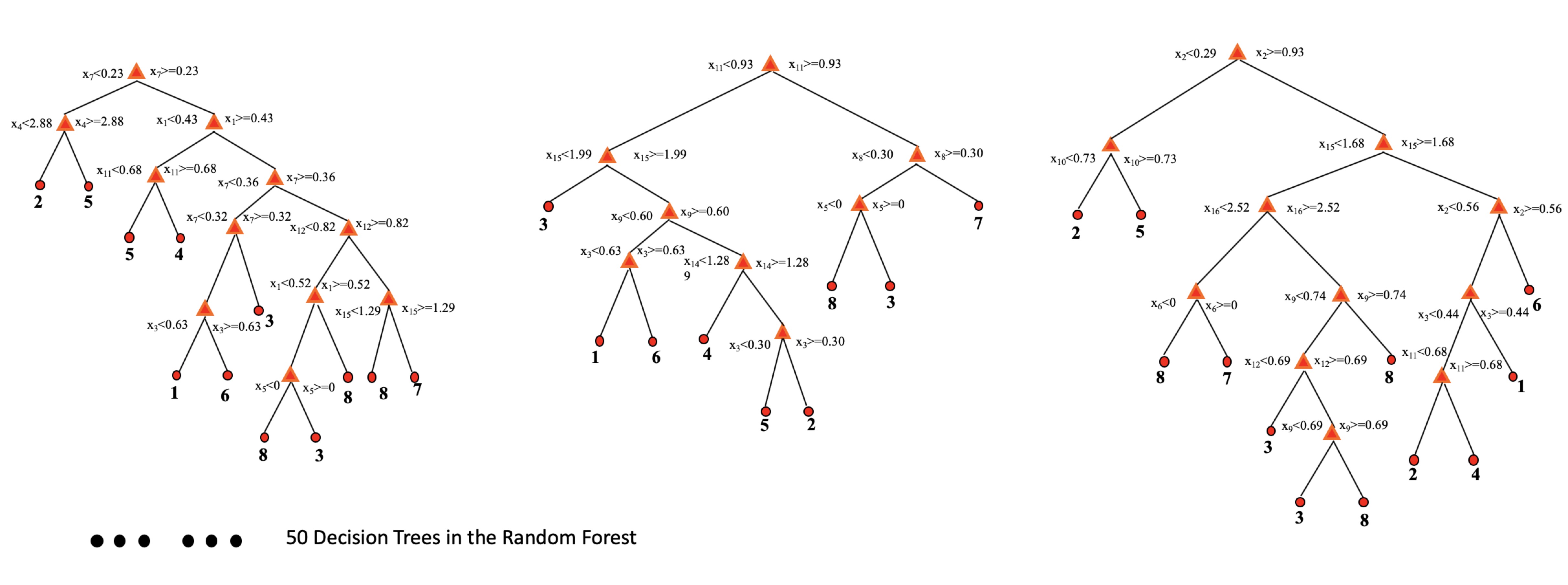}
\end{center}
   \caption{part of trained result of random forest}
\label{fig:short}
\end{figure*}

\section{Experimental Results}
In order to validate the effectiveness of the proposed algorithm, this section includes four parts. To validate effectiveness of curvature as a signature of action itself, we firstly calculate curvatures on a relatively simpler dataset, which doesn't have much noise or disturbance from background. Features are generated from the curvatures and fed into random forest to obtain the classification result. Robust analysis against flipping is carried out. Then, to further validate our algorithm, Mask R-CNN is customized as a data-augmentation process. And comparison experiment among several different methods are carried out.

\begin{figure*}
\begin{center}
\includegraphics[width=1\linewidth]{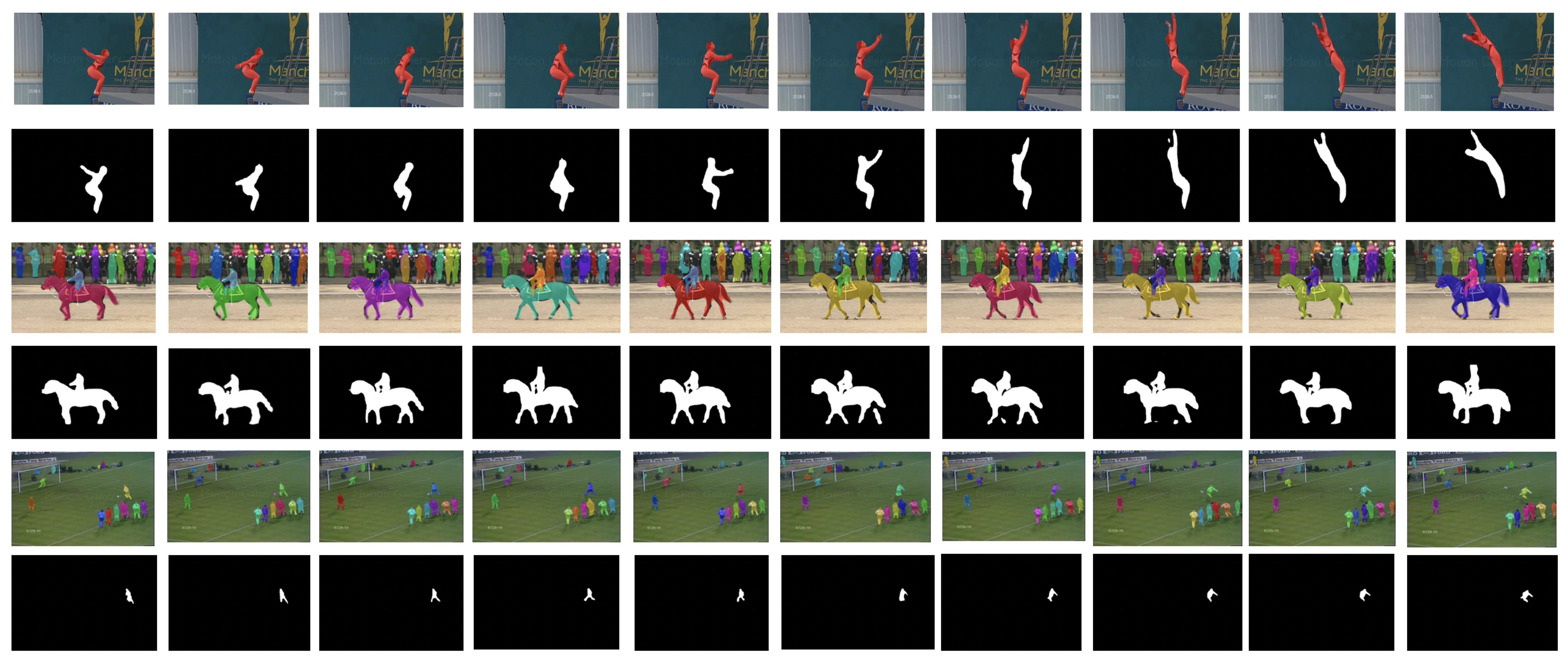}
\end{center}
   \caption{Transfer learning result of Mask R-CNN to our task}
\label{fig:short}
\end{figure*}

\begin{figure*}
\begin{center}
\includegraphics[width=1\linewidth]{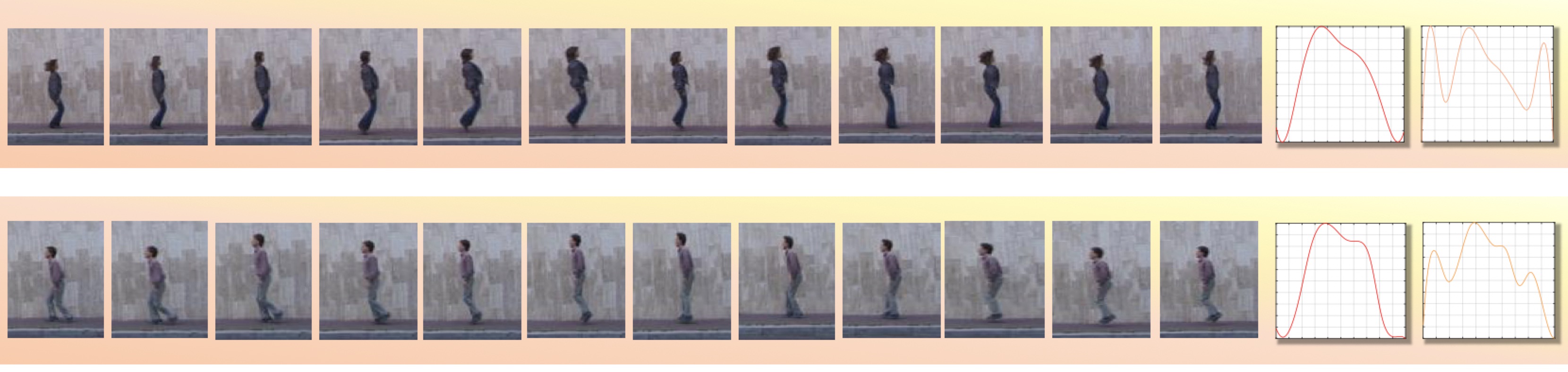}
\end{center}
   \caption{Comparison of curvature when filpping happens}
\label{fig:short}
\end{figure*}

\subsection{Validation of the Curvature Description}
This part aims at validating effectiveness of curvature as a signature of action recognition. The experiments are carried out on Weizmann dataset, whose results are shown in Fig.2.  First curvatures (usually refered to as curvature) and second curvatures (usually refered to as torsion) of eight different actions are plotted  and analyzed. In Fig2, eight classes of actions are considered, respectively waving with one hand, waving with two hand, jumping side, jumping up, bending, walking, siding, and skipping. The first curvatures are plotted according to equation (4), with respect to universal time scale. It can be observed that the first curvatures of each particular action follow certain particular pattern, and the first curvatures of different actions show different patterns. Second curvatures are calculated according to equation (5).\\
\indent Theoretically speaking, we could calculate as high order of curvatures as we want. In practice, however, derivative calculation comes with computational error. And as the order of derivative goes higher, this error goes larger. In order to make the calculation of derivative as precise as possible, we used finite difference method to calculate the derivatives. And we found out that second curvature, which involves third derivative as the highest order, has acceptable scale of error. And including first curvature and second curvature is enough for the classification task. So there is no need to involve curvatures of higher orders here.\\
\indent Since the curvatures are obtained, statistical features could be generated from them. Experimental results carried out on Weizmann dataset are shown in TABLE1.\\

\begin{table}
\begin{center}
\begin{tabular}{|p{5cm}|p{2cm}|}
\hline
Method & Performance \\
\hline\hline
PBMS\cite{Authors034} & 0.870\\
CSTIP\cite{Authors035} & 0.967\\
Vanilla 1st Curvature & 0.850 \\
Vanilla 1st, 2nd Curvature & \textbf{0.950}\\
\hline
\end{tabular}
\end{center}
\caption{Comparison to other methods on on Weizmann dataset.}
\end{table}

\begin{table}
\begin{center}
\begin{tabular}{|p{5cm}|p{2cm}|}
\hline
Method & Performance \\
\hline\hline
MACH\cite{Authors036} & 0.692\\
LTP\cite{Authors037} & 0.793\\
DFCM\cite{Authors038}  & 0.837\\
HOG+FV\cite{Authors018} & 0.850\\
Mask R-CNN+1st Curvature & 0.759 \\
Mask R-CNN+1st, 2nd Curvature & \textbf{0.870}\\
\hline
\end{tabular}
\end{center}
\caption{Comparison to other methods on on UCF Sports dataset.}
\end{table}

\subsection{Robustness Analysis}
Another advantage of the proposed curvature signature lies in the fact that it is robust against flipping. This is because the pattern extracted by our algorithm doesn't depend on matching the location of pixels between frames. It could be observed from Fig 5 that symmetric actions result in similar pattern in curvatures, with slight intraclass difference. Thus the curvature as a signature is robust against flipping.

\subsection{Data Augmentation}
With previous experiments, we proved the effectiveness of curvature as a signature of action. Because background of Weizmann dataset is basically always the same wall not moving, simple matrix subtraction could be carried out to remove the back ground. However, for more complex videos, which are know as videos in the wild, fancier background removal algorithm need to be applied. In order to carry out our algorithm in wild videos, we customize Mask R-CNN\cite{Authors032} as a data augmentation process. Because our algorithm is a few-shot learning, experiments are carried out on a smaller sample of the well-known UCF101 dataset\cite{Authors033}, namely UCF Soorts dataset. Fig4 shows the data augmentation result with MASK R-CNN on consecutive frames of a video. It should be noted that MASK R-CNN was trained on COCO dataset which consists of still images. When it comes to videos, motion would introduce extra noises. Background clustering and poor lighting are also potential reasons that could make transfer learning fail. We fine-tuned the last layer of MASK R-CNN in order to accomodate to our task.

\subsection{Comparison Experiment}
We compare the performances of published methods on Weizmann dataset as well as UCF Sports dataset in TABLE I and TABLE II respectively. It could be observed that first derivative is an effective signature of action, and performance is better when both first and second curvatures are used. We focused most of our narrative on the modeling. It should be noted that the proposed signature of curvature is a universal feature. Classifier is not limited to Random Forest. Other state-of-art classification architectures such as CNN or LSTM could also be incorporated, and multiple features could be incorporated in order to obtain higher precision in classification task. 

\begin{figure}[thpb]
      \centering
       \includegraphics[width=0.5\textwidth]{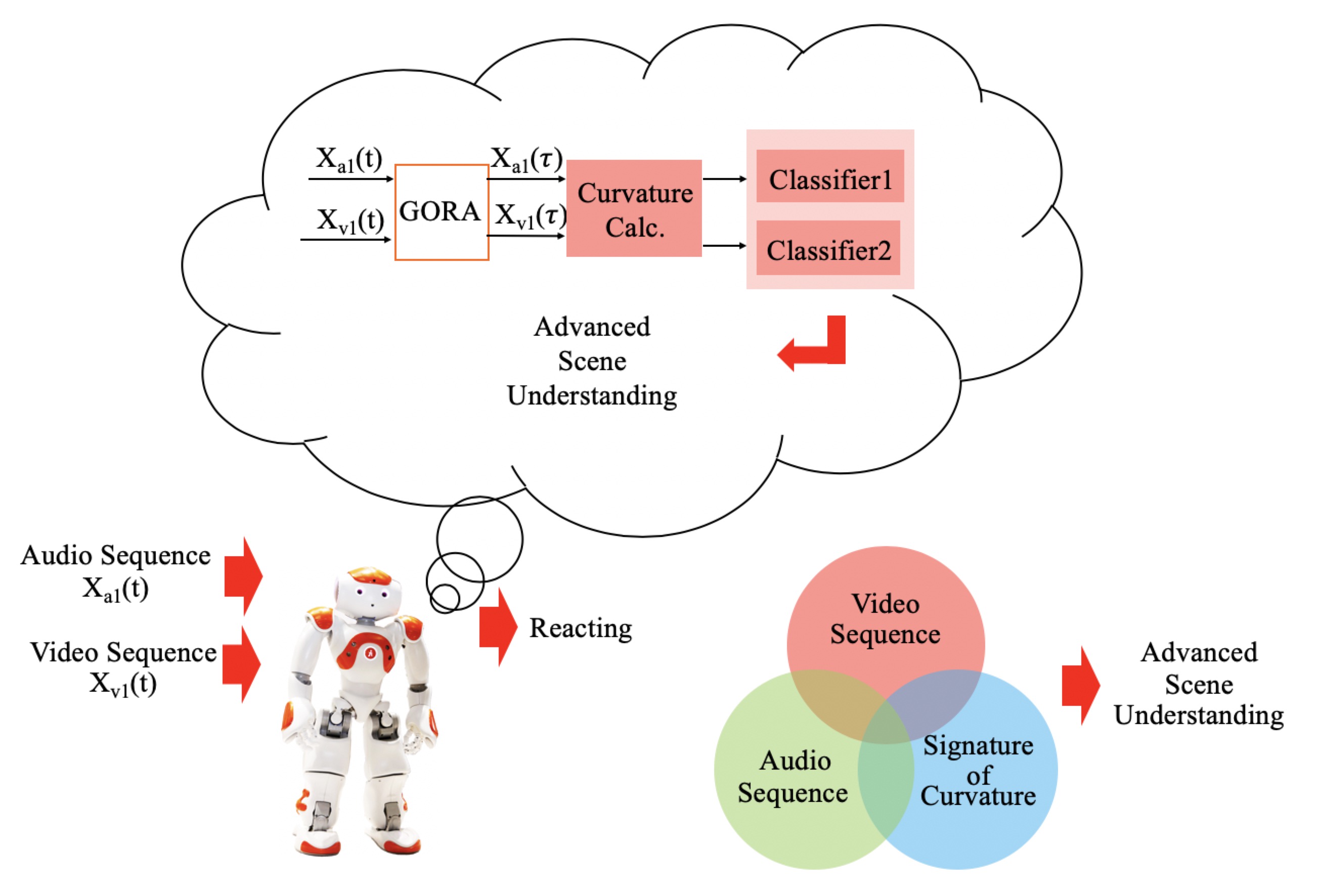}
      \caption{The goal of our future research}
      \label{figurelabel}
  \end{figure}

\section{Conclusions}

We proposed a new description of actions based on the curvatures of sequences. This is in contrast to most previous learning-based algorithms, which use edges as description for classification. Building upon the curvature description, features are extracted by calculating statistical indexes. Based on such features, we demonstrate our algorithm in the global time scale instead of a few frames. And it should be noted that the proposed method is proved to be effective in but not limited to video sequences. It could also be accommodated for inputs of audio sequence such as language or music. In the future, we would like to look into details of how to customize this algorithm to a fusion of video and audio sequences. As shown in Fig.6, we believe the fusion of video sequence, audio sequence, and signature of curvature would lead to promising performance of advanced scene understanding, which would open a new prospect for smart service robot\\
\textbf{Acknowledgements}\\
This work is supported by Office of Naval research Award (ONR) N00014-17- 1-2142. The authors appreiciate Ting Da for resizing the masks. We also appreciate Mengdi Xu, Thomas Mitchel, Weixiao Liu, and Sipu Ruan for discussion.

{\small
\bibliographystyle{ieee_fullname}
\bibliography{egbib}
}

\end{document}